\def\paperID{4716}
\def\confName{ICCV}
\def\confYear{2025}

\def\paperTitle{Keyframe-oriented Vision Token Pruning: Enhancing Efficiency of Large Vision Language Models on Long-Form Video Processing}

\def\authorBlock{
    Yudong Liu$^{1}$ \qquad
    Jingwei Sun$^{1}$ \qquad
    Yueqian Lin$^{1}$ \qquad
    Jianyi Zhang$^{1}$ \qquad
    Jingyang Zhang$^{1}$ \\
    Ming Yin$^{1}$ \qquad
    Qinsi Wang$^{1}$ \qquad
    Guangzhi Su$^{2}$ \qquad
    Hai Li$^{1}$ \qquad
    Yiran Chen$^{1}$ \\
    $^{1}$Duke University \\
    $^{2}$Duke Kunshan University \\
    {\tt\small \{first.last\}@duke.edu}
}

\newif\ifreview 
\newif\ifarxiv \newcommand{\arxiv}{\arxivtrue}
\newif\ifcamera 
\newif\ifrebuttal 

\arxiv 

\pdfoutput=1
\documentclass[10pt,twocolumn,letterpaper]{article}
\ifreview \usepackage[review]{cvpr} \fi
\ifarxiv \usepackage[pagenumbers]{cvpr} \fi
\ifrebuttal \usepackage[rebuttal]{cvpr} \fi
\ifcamera \usepackage{cvpr} \fi

\usepackage{graphicx}	
\usepackage{amsmath}	
\usepackage{amssymb}	
\usepackage{booktabs}
\usepackage{times}
\usepackage{microtype}
\usepackage{epsfig}
\usepackage{caption}
\usepackage{float}
\usepackage{placeins}
\usepackage{color, colortbl}
\usepackage{stfloats}
\usepackage{enumitem}
\usepackage{tabularx}
\usepackage{xstring}
\usepackage{multirow}
\usepackage{xspace}
\usepackage{url}
\usepackage{subcaption}
\usepackage{xcolor}
\usepackage{listings}
\usepackage{tcolorbox}
\usepackage{xcolor}
\usepackage[T1]{fontenc}
\usepackage{lmodern}

\usepackage[subtle,margins=normal,leading=normal]{savetrees}

\usepackage[hang,flushmargin]{footmisc}

\ifcamera \usepackage[accsupp]{axessibility} \fi

% ...

\ifarxiv  \fi

\newcommand{\R}[1]{{%
    \textbf{%
        \ifstrequal{#1}{1}{\textcolor{red}{R#1}}{%
        \ifstrequal{#1}{2}{\textcolor{blue}{R#1}}{%
        \ifstrequal{#1}{3}{\textcolor{magenta}{R#1}}{%
        \ifstrequal{#1}{4}{\textcolor{teal}{R#1}}{%
                           \textcolor{cyan}{R#1}%
        }}}}%
    }%
}}

\usepackage{xr-hyper}

\makeatletter
\newcommand*{\addFileDependency}[1]{
  \typeout{(#1)}
  \@addtofilelist{#1}
  \IfFileExists{#1}{}{\typeout{No file #1.}}
}

\makeatother
\newcommand*{\myexternaldocument}[1]{
    \externaldocument{#1}
    \addFileDependency{#1.tex}
    \addFileDependency{#1.aux}
}

\definecolor{cvprblue}{rgb}{0.21,0.49,0.74}
\usepackage[pagebackref,breaklinks,colorlinks,allcolors=cvprblue]{hyperref}
\usepackage[capitalize]{cleveref}
\crefname{section}{Sec.}{Secs.}
\crefname{table}{Table}{Tables}
\crefname{figure}{Fig.}{Figs.}

\ifarxiv \crefname{appendix}{App.}{Apps.}
\else \crefname{appendix}{Suppl.}{Suppls.} \fi

\unless\ifarxiv \myexternaldocument{_supplementary} \fi

\begin{document}
\title{\paperTitle}
\author{\authorBlock}
\maketitle

\begin{abstract}
Vision language models (VLMs) demonstrate strong capabilities in jointly processing visual and textual data. However, they often incur substantial computational overhead due to redundant visual information, particularly in long-form video scenarios. Existing approaches predominantly focus on either \textbf{vision token pruning}, which may overlook spatio-temporal dependencies, or \textbf{ keyframe selection}, which identifies informative frames but discards others, thus disrupting contextual continuity. In this work, we propose \textbf{KVTP (Keyframe-oriented Vision Token Pruning)}, a novel framework that overcomes the drawbacks of token pruning and keyframe selection. By adaptively assigning pruning rates based on frame relevance to the query, KVTP effectively retains essential contextual information while significantly reducing redundant computation. To thoroughly evaluate the long-form video understanding capacities of VLMs, we curated and reorganized subsets from seven datasets into a unified benchmark that highlights real-world scenarios with sparse but crucial events. Our experiments with VLMs of various scales show that KVTP can reduce token usage by $80\%$ without compromising spatiotemporal and contextual consistency, significantly cutting computation while maintaining the performance. These results demonstrate our approach's effectiveness in efficient long-video processing, facilitating more scalable VLM deployment. The code for this paper is available at \href{<https://github.com/1999Lyd/KVTP.git>} {\texttt{GitHub}}.

\end{abstract}

\section{Introduction}
\label{sec:intro}

One variant of multimodal large language models (MLLMs), the vision-language model (VLM) \cite{zhang2024llavanext-video, lin2023video, damonlpsg2023videollama, 2023videochat}, has demonstrated exceptional capabilities in tackling challenging video-language reasoning tasks. A typical VLM workflow involves sampling a video into a sequence of frames, encoding the frames into a long sequence of vision tokens using a vision encoder, and concatenating these vision tokens with query tokens before feeding the combined sequence into the backbone LLM for inference.

\begin{figure}[tp]
    \centering
    \includegraphics[width=0.9\linewidth]{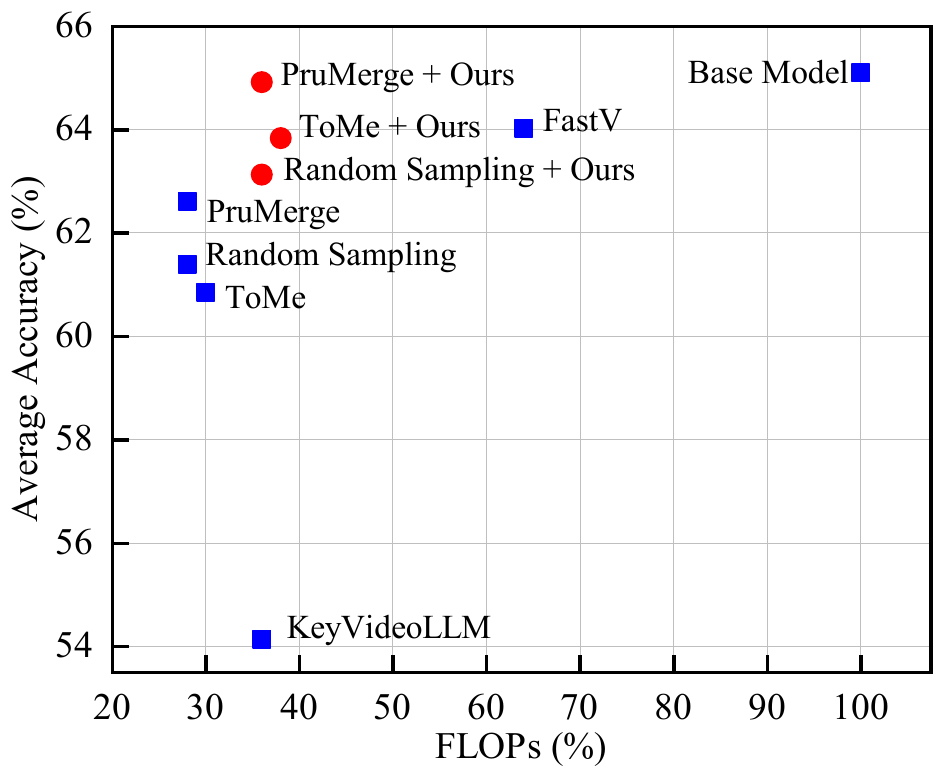}
    \caption{Performance vs. FLOPs across different methods. Results are averaged over SparseKV-QA, which comprises three subsets from VideoMME, EgoSchema, and NeXT-QA.}
    \label{fig:performance}
    \vspace{-4mm}
\end{figure}
\begin{figure}[tp]
    \centering
    \includegraphics[width=\linewidth]{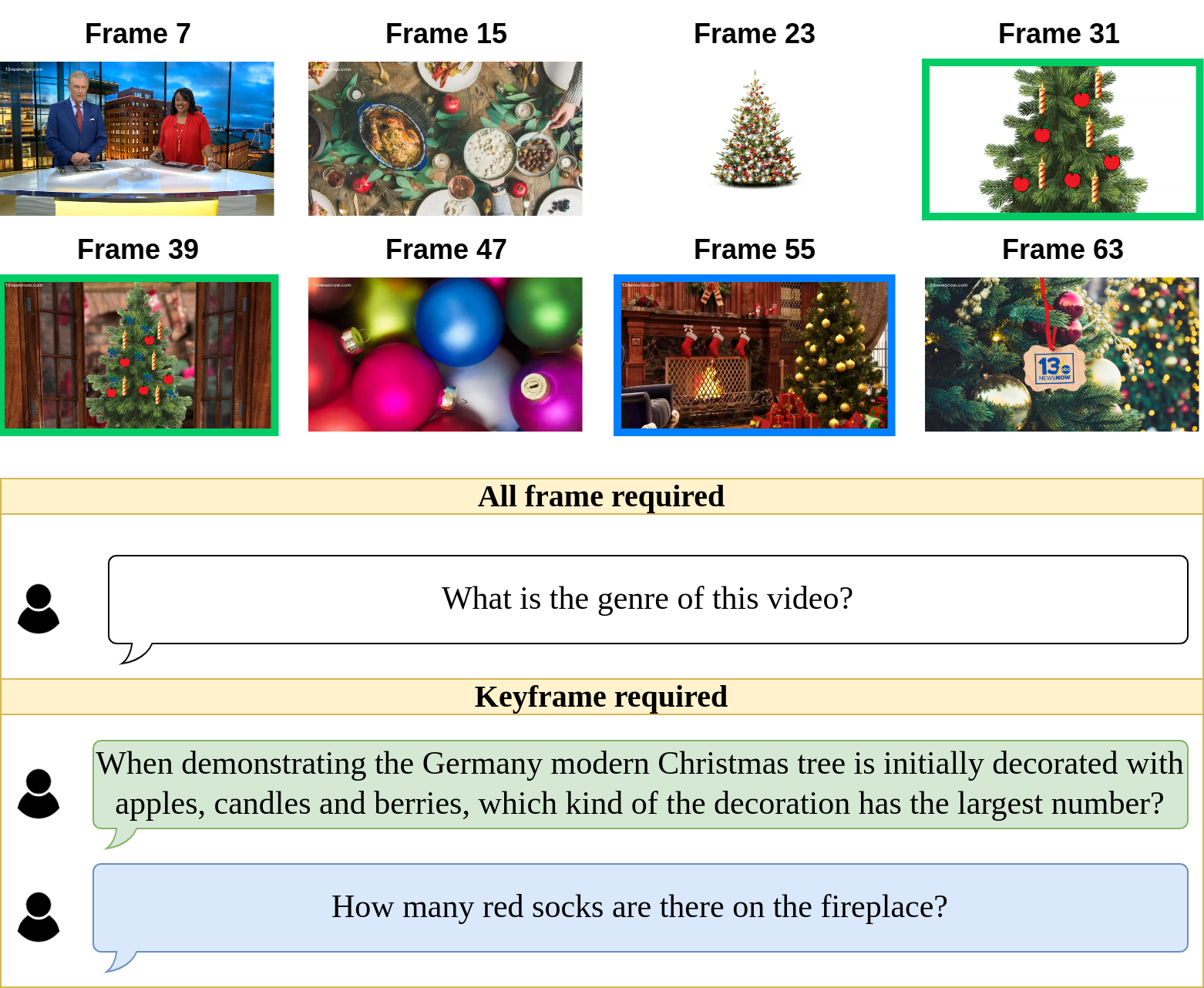}
    \caption{A typical sample from the VideoMME dataset. While the first question requires all frames for a comprehensive answer, the second and third questions focus on only a small subset of frames (highlighted in colored borders). These latter questions align with our target scenario, where identifying the relevant frames from the entire video is paramount.}
    \label{fig:data_sample}
    \vspace{-4mm}
\end{figure}

For most state-of-the-art (SOTA) pipelines \cite{zhang2024llavanext-video, lin2023video, damonlpsg2023videollama, 2023videochat}, the entire sequence of video tokens (which can exceed 10,000 tokens for videos longer than 1 minute) is processed by the LLM backbone for every conversation. This approach introduces significant redundancy and computational overhead, posing a challenge for the efficient deployment of VLMs. As illustrated in Figure~\ref{fig:data_sample}, which depicts a typical conversation from the VideoMME \cite{fu2024video} dataset. While certain queries require the full sequence of video frames for accurate responses, many real-world applications, such as surveillance systems, video analytics, and content summarization, often involve queries focusing on only a small, specific part of the long video. In such cases, processing the entire video token sequence in every conversational turn is highly inefficient, especially given the quadratic computational complexity of the attention mechanism with respect to token sequence length.

Existing works \cite{choi2024vid,rao2021dynamicvit,shang2024LLaVA-PruMerge,chen2024image,lin2024speechprune,kong2022spvit,wei2023joint,Park2024TooMF,wang2024videotree,keyvideo,koala,han2024videoespresso} primarily take two approaches for this problem: vision token pruning and keyframe selection. Most vision token pruning methods were originally developed for image-language tasks and primarily focus on reducing spatial redundancy within images. When applied to long-form video tasks, these methods process each frame independently, failing to capture the inter-frame temporal connections. On the other hand, conventional keyframe selection methods typically employ a \textit{hard selection} strategy, where frames deemed irrelevant are entirely discarded once they fall below a predefined threshold. However, even these less relevant frames may contain crucial tokens necessary for maintaining the logical and contextual structure of the video, which is crucial for video reasoning task. Removing them can result in significant information loss, ultimately degrading the performance of video-language models. 

To address these challenges, we propose \textbf{KVTP (Keyframe-oriented Vision Token Pruning)}, which employs \textit{soft selection} to retain a small fraction of tokens from less relevant frames to preserve essential cues for high-level reasoning. This design effectively reduces redundancy while maintaining temporal and contextual coherence, as illustrated in Figure~\ref{fig:example} (more qualitative results in Appendix~\ref{sec:quali}) and further validated by experiments in Section~\ref{sec:experiments}. Our method adaptively identifies frame importance based on the query and video context. This pipeline features a novel frame relevance predictor that adaptively determines query-frame relevance during inference. Once relevance scores are assigned to all frames, a soft frame selection mechanism translates these scores into frame-specific pruning rates, retaining more tokens from keyframes while reducing tokens from less relevant frames. This approach reduces token length significantly while preserving the temporal and contextual structure of the original video, with detailed formulations provided in section~\ref{sec:method}.

To benchmark the performance of all efficiency vision token compression methods in the target problem setting, we designed a pipeline to create a benchmark dataset, named \textbf{SparseKV-QA} (Sparse Key Video Question Answering), featuring long videos with sparse key information. This dataset is constructed as a subset of seven video QA datasets \cite{2023ego,xiao2021next,fu2024video, yu2019activityqa,patraucean2023perception,Maaz2023VideoChatGPT,li2024mvbench}, where long videos containing sparse keyframes are selected based on predefined criteria. We introduce this new dataset to benchmark our target setting, as no existing open-source benchmark dataset specifically focuses on long-form video reasoning with sparse key information. As shown in Figure~\ref{fig:data_sample}, it is crucial to distinguish between the cases where keyframe prediction is beneficial and those where the entire video is necessary to answer a query. This benchmark would be essential for providing a structured and meaningful way to evaluate vision token compression methods under this simulated real-world scenario.

\begin{figure}[tp]
    \centering
    \includegraphics[width=\linewidth]{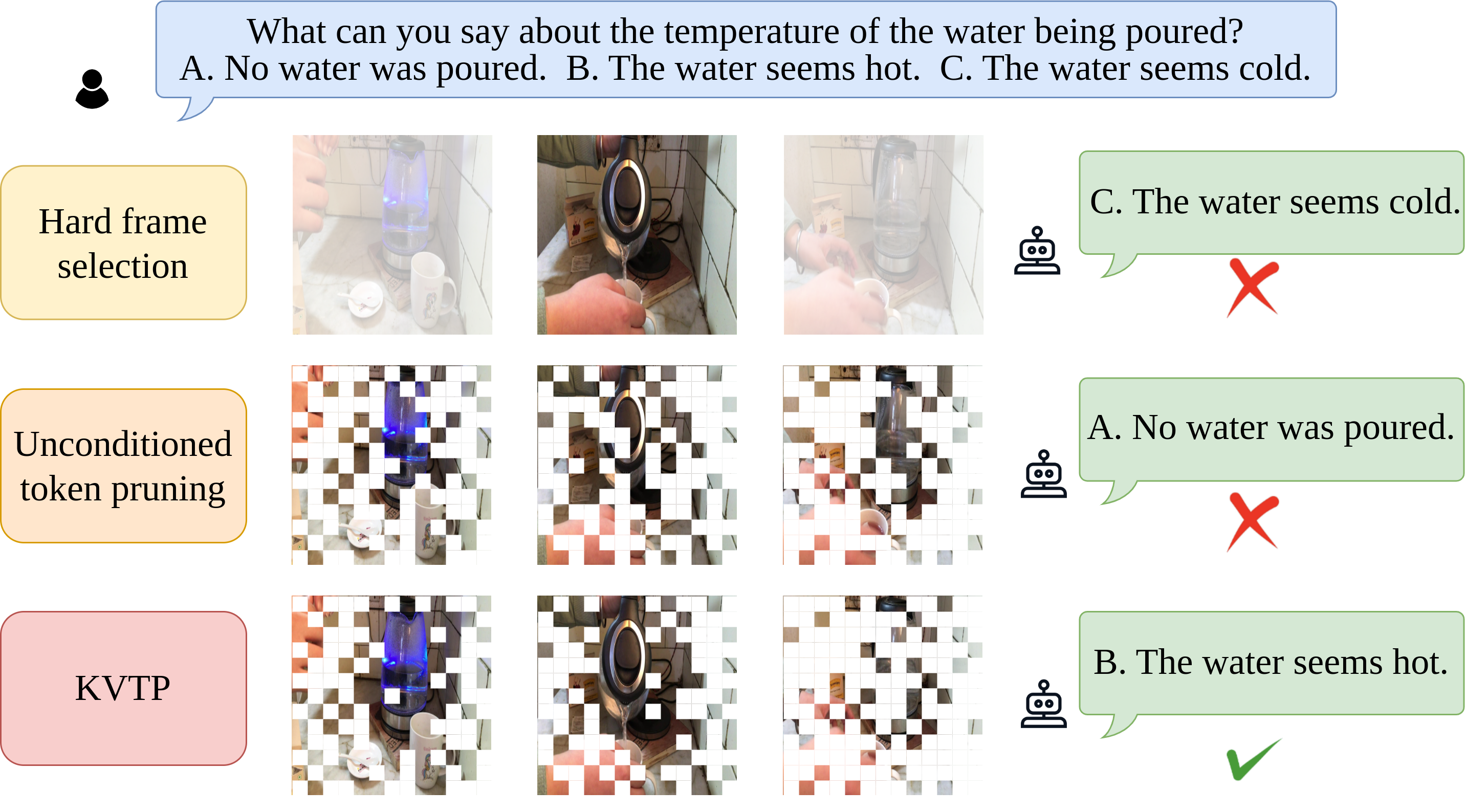}
    \caption{A representative comparison between 3 types of efficiency pipeline from the LLaVA-Video-7B benchmark on Perception test. The top row illustrates \textbf{hard frame selection}, where only the most query-relevant frames are retained, causing the model to miss the events leading up to the water being poured, resulting in an incorrect answer. The middle row demonstrates \textbf{unconditioned token pruning} across frames, where excessive information loss in key frames prevents the model from determining whether the water was poured. The bottom row showcases \textbf{keyframe-oriented token pruning}, where most tokens from key frames are preserved while retaining some tokens from other frames to maintain contextual and temporal coherence, enabling the VLM to produce the correct answer.}
    \label{fig:example}
    \vspace{-4mm}
\end{figure}
Our method bridges the gap between keyframe selection and image token pruning by converting frame relevance scores into frame-wise pruning rates. This \textit{keyframe-oriented vision token pruning} paradigm unifies coarse-grained frame selection with fine-grained token-level pruning, enabling efficient processing of long-form videos while preserving model performance. To summarize, our contributions are as follows:
\begin{enumerate}
    \item We identify and formalize a prevalent limitation in long-form video question answering tasks and develop a cost-effective pipeline to construct a new dataset using existing long-video benchmarks.
    \item We propose a novel approach to fine-tune vision encoders, yielding a plug-and-play query-frame relevance predictor tailored for long-video question-answering tasks.
    \item Our method adaptively assigns pruning rates to each frame, a key factor in many existing vision token pruning pipelines, and serves as a plug-and-play module that consistently improves token pruning performance on long-video benchmarks. Remarkably, as shown in Figure~\ref{fig:performance}, it can prune up to 80\% of video tokens and result in 64\% FLOPs reduction while preserving the original performance.
\end{enumerate}

\section{Related Work}
\label{sec:related}

\subsection{Multimodal Large Language Models (MLLMs)}

The emergence of multimodal large language models (MLLMs) began with works such as LLaVA and InstructBLIP \cite{liu2023llava,2023instructblip}, which primarily focused on image language tasks. By incorporating a pre-trained vision encoder such as CLIP \cite{2021clip} and an adapter (e.g., MLP or Q-former) to project vision tokens into the text token space, these models enable backbone LLMs to understand vision information in the semantic domain. After fine-tuning with carefully crafted instruction-tuning data, these models demonstrate strong cross-modal reasoning abilities.

Recently, research has shifted towards video-language tasks \cite{zhang2024llavanext-video, lin2023video, damonlpsg2023videollama, 2023videochat,2023ego,xiao2021next,fu2024video,chen2023sharegpt4v,patraucean2023perception,ren2024timechat,chai2024auroracap}. For instance, LLaVA-NeXT \cite{zhang2024llavanext-video} samples videos into sequences of frames and performs video-text inference akin to multi-image text inference. However, since video-language models typically process each frame independently within vision encoders, they often struggle to capture temporal relationships. To address this, LLaVA-NeXT introduces a temporal instruction as a preprompt to inform the LLM of the video sampling process, while VideoLLaVA \cite{lin2023video} employs a learnable temporal embedding within the vision encoder. However, this approach imposes a constraint where the number of sampled frames during inference must match the training stage.

\subsection{vision Token Pruning}

The large number of vision tokens is a primary bottleneck for efficient video language modeling. Recent works \cite{choi2024vid,rao2021dynamicvit,shang2024LLaVA-PruMerge,chen2024image,lin2024speechprune,kong2022spvit,wei2023joint} propose advanced vision token pruning algorithms to significantly reduce token counts while maintaining competitive performance. These methods offer diverse metrics to evaluate token importance:

\begin{itemize}
    \item \textbf{Vid-TLDR:} Proposes a saliency score based on the entropy of attention within a single image, where a higher entropy implies more informative tokens.
    \item \textbf{DynamicVit:} Introduces a learnable token importance predictor module leveraging global and local information within the image to select key tokens.
    \item \textbf{PruMerge:} Selects important tokens based on the final-layer attention scores between the class token and other tokens in ViT.
    \item \textbf{FastV:} Aggregates attention scores across selected layers within the LLM backbone to identify significant tokens.
\end{itemize}

Although these methods and similar approaches in other modalities excel in their respective domains, their direct application to long video-language tasks, treating each frame separately without considering interframe context information and using uniform hyperparameters for each frame, results in suboptimal outcomes due to frame-wise redundancy in videos with sparse key information. Intuitively, keyframe identification can serve as a crucial step in optimizing state-of-the-art vision token pruning pipelines in such cases.

\subsection{Keyframe Identification}

In video-language tasks, processing a large number of frames leads to significant computational costs. Given the inherent frame-wise redundancy in videos, keyframe selection algorithms \cite{Park2024TooMF,wang2024videotree,keyvideo,koala,han2024videoespresso} have emerged as an effective solution. Keyframe selection typically involves two steps:

\textbf{Step 1: Video Segmentation or Frame Clustering.}  
Video frames often exhibit temporal redundancy, and individual frames carry limited information. Methods like LVNet cluster frames based on the similarity of latent frame embeddings, while VideoTree applies K-means clustering. KeyVideoLLM, and Koala perform chronological segmentation.

\textbf{Step 2: Context-Aware Keyframe Identification.}  
Context awareness is often derived from queries. LVNet reconstructs queries into multiple keywords to filter noise and computes similarities between keywords and segment embeddings. VideoTree generates short captions for each cluster using GPT and compares queries with these captions to assign relevance scores. KeyVideoLLM follows a similar approach to ours, utilizing pretrained vision and text encoders to compute cosine similarity as a frame relevance metric.

\section{SparseKV-QA Construction}
\label{sec:data}

\begin{figure*}[t]
    \centering
    \includegraphics[width=0.85\linewidth]{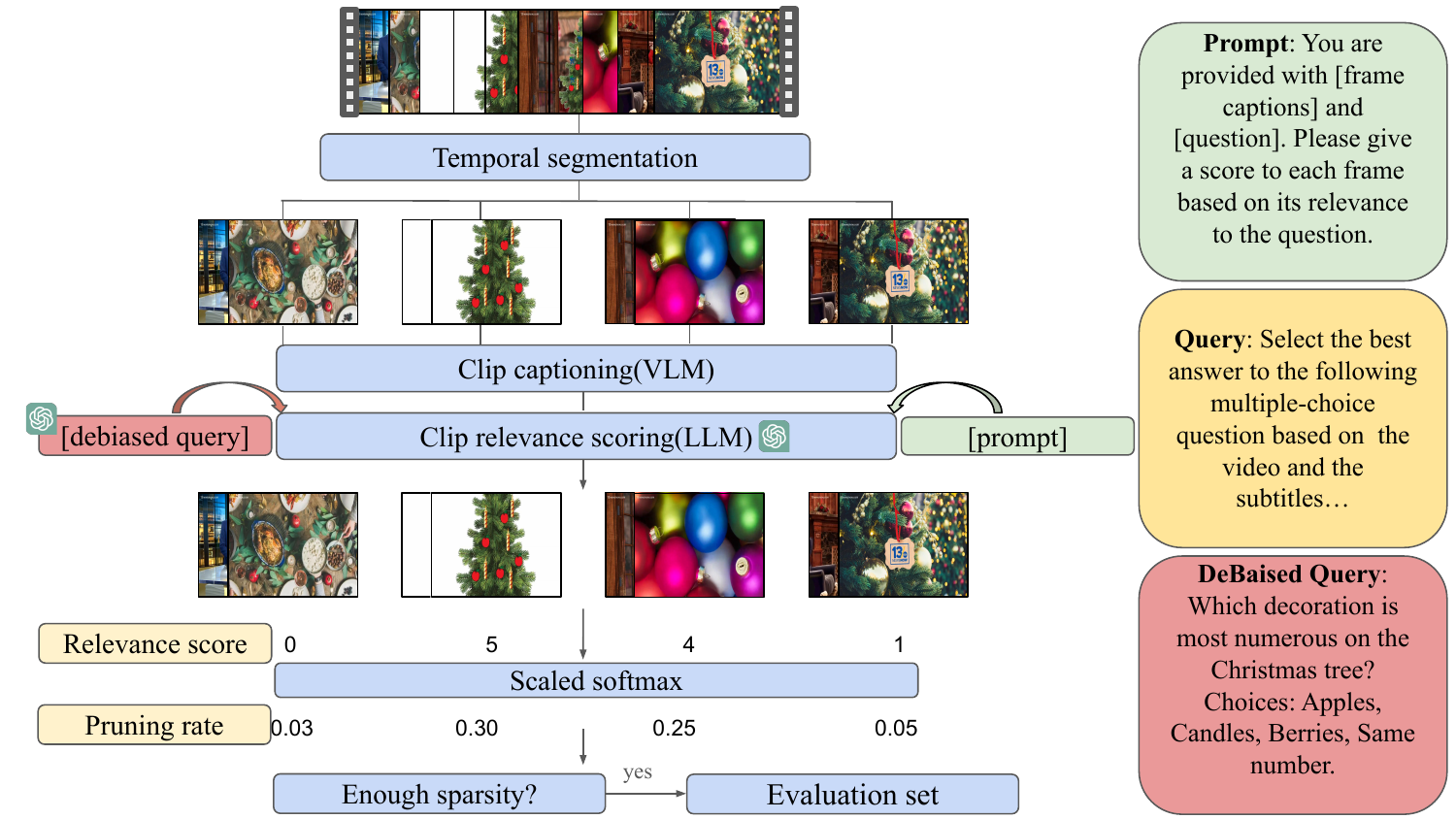}
    \caption{Overview of the proposed data augmentation framework, applied to the VideoMME, EgoSchema, and NeXT-QA datasets. This process enhances each video with clip-level captions, de-biased queries, and query-frame relevance scores. The dataset is then divided into evaluation and training sets based on keyframe sparsity, as determined by the relevance scores.}
    \label{fig:data}
    \vspace{-2mm}
\end{figure*}

An overview of the task formulation and benchmark creation pipeline is shown in Figure~\ref{fig:data}. Given a long video $V$, the video is first sampled into $N$ frames using a uniform sampling strategy, following the LLaVA-Video pipeline. LLaVA-Video set $N$ to 64. To handle longer videos, we increase the maximum value of $N$ from 64 to 128, ensuring compatibility with long-video tasks. After sampling, the $N$ frames are grouped into clips in chronological order, where each clip contains $n$ frames. We set $n=8$, resulting in $k = N/n$ clips per video, denoted as $c_1, c_2, \dots, c_k$.

To reduce caption costs, we use the LLaVA-NeXT model as the captioner to generate captions for each clip:
\[
t_i = F_\text{cap}(c_i), \quad i = 1, \dots, k.
\]
These captions, combined with the query $Q$, are fed into a large language model (LLM) to label relevance scores for each frame:
\[
S = (s_1, s_2, \dots, s_N) = F_\text{score}((t_1, t_2, \dots, t_k), Q).
\]
In this step, we utilize GPT-4o to ensure high-quality labeling. However, this straightforward text-only task can also be performed using a simpler language model, such as BERT~\cite{devlin2018bert}, if budget constraints are a concern.

As video reasoning benchmarks often contain lengthy and noisy queries, potentially degraded by additional system prompts or irrelevant context, we introduce a query refinement step inspired by~\cite{vmr}. Before relevance scoring, we employ GPT-4o to shorten and debias the query, ensuring it retains essential information while removing unnecessary noise. This preprocessing step enhances the quality of the frame relevance scores and benefits the predictor fine-tuning pipeline, as the maximum query length for pretrained SigLIP is 64 tokens, constrained by its pretrained positional embeddings.

Next, a softmax function is applied to the relevance scores to measure the probability of maintaining the corresponding frames, which is then converted into pruning rates by scaling it with sequence length and a predefined average pruning rate:
\begin{equation}
    R = (r_1, r_2, \dots, r_N) = \alpha N \cdot \text{softmax}(s_1, \dots, s_N),
    \label{eq:pruning_rate}
\end{equation}
where $\alpha$ is a hyperparameter controlling the expected pruning rate. As higher maximum pruning rates indicate higher sparsity of keyframes, we filter out videos that do not meet the following criteria:
\begin{itemize}
    \item $\max(r_1, \dots, r_N) < \beta \alpha$ (not sparse enough),
    \item $N < 64$ (not long enough).
\end{itemize}
We construct SparseKV-QA by filtering videos from seven existing video-language datasets: VideoMME, EgoSchema, NextQA, ActivityNetQA \cite{yu2019activityqa}, Perception Test \cite{patraucean2023perception}, VideoChatGPT \cite{Maaz2023VideoChatGPT}, and MVBench \cite{li2024mvbench}. Dataset statistics are presented in Table~\ref{tab:train_set}. 

\begin{table}[h]\small
\footnotesize
    \centering
    \begin{tabular}{lc c c}
        \toprule
        Dataset & Size & Avg. Length & Sparsity \\
        \midrule
        Full Dataset  & 50,812 & 245s & 1.95 \\
        SparseKV-QA & 20,050 & 451s & 3.56 \\
        \bottomrule
    \end{tabular}
    \caption{Dataset statistics for the full dataset and SparseKV-QA. Keyframe sparsity is measured as the ratio of the maximum pruning rate to the average pruning rate within a video.}
    \label{tab:train_set}
    \vspace{-3mm}
\end{table}

\section{Method}
\label{sec:method}

\begin{figure*}[t]
    \centering
    \includegraphics[width=0.85\linewidth]{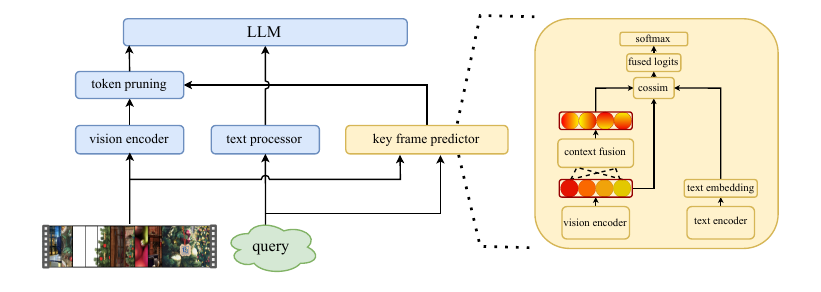}
    \caption{Overview of the proposed KVTP framework. A keyframe predictor module is integrated to guide the token pruning process. A context fusion head is incorporated above the vision encoder, aggregating contextual information from both local clips and the global video. The fused logits, which encode both context and query-frame relevance information, are then converted into pruning rates as the final output.}
    \label{fig:framework}
    \vspace{-2mm}
\end{figure*}

Our method is designed to optimize state-of-the-art vision token pruning pipelines for challenging long-video tasks with sparse key information. By adaptively measuring the query-frame relevance during inference and converting this relevance into frame-wise pruning rates, we aim to achieve efficient video-language model (VLM) inference. An overview of our framework is shown in Figure~\ref{fig:framework}. The method consists of two steps: (1) fine-tuning the query-frame relevance predictor using augmented data and a crafted objective function, and (2) keyframe-oriented vision token pruning. These steps are detailed in the following subsections.

\subsection{Query-Frame Relevance Predictor Fine-Tuning}
\label{sec:relevance_predictor_finetuning}

The foundation of our query-frame relevance predictor is the pretrained vision encoder SigLIP \cite{2023siglip}, as its original contrastive learning objective aligns closely with our goal of identifying image-text relevance. In the context of long-video tasks, given a sequence of frames $f_1, f_2, \dots, f_N$ and a query $Q$, we can apply the SigLIP model to generate embeddings for them:
\[
E = \{e_1, e_2, \dots, e_N\} = F_\text{img}(f_1, f_2, \dots, f_N), \]
\[ e_q = F_\text{text}(Q).
\]
The relevance logits $L$ between the frames and query are computed as:
\begin{equation}
    L = a \cdot \text{cos\_sim}(E, e_q) + b,
\end{equation}
where $a$ and $b$ are learnable scaling and bias parameters. The frame-wise relevance distribution are then derived using:
\begin{equation}
    P = (p_1, p_2, \dots, p_N) = \text{softmax}(L),
\end{equation}

A naive approach to construct the objective function would be to directly apply the contrastive loss between the logits generated by the encoder and the ground truth, where the ground truth label $s_1, s_2, \dots, s_N$ can be constructed by the pipeline introduced in Section~\ref{sec:data}. SigLIP optimizes contrastive loss by assigning positive weights to the logits of correlated image-text pairs while applying negative weights to unrelated pairs before aggregating the logits. Following this approach, we first normalize the ground truth relevance scores by subtracting their mean value:

\begin{equation}
    \hat{S} = (s_1, s_2, \dots, s_N) - \frac{1}{N} \sum_{i=1}^{N} s_i.
\end{equation}

Next, the contrastive loss is computed using an element-wise product:

\begin{equation}
    O = -\sum \log \sigma (L \odot \hat{S}),
\end{equation}

where $\sigma$ denotes the sigmoid function and $\odot$ represents the element-wise product.

However, this formulation treats the frame sequence as a batch of independent images and ignores the temporal and contextual connections between frames. To address this limitation, we introduce a context fusion head that integrates local and global contextual information.

\paragraph{Context Fusion Head}
The context fusion head consists of two components:

\textbf{Local Context Fusion Head:} This module performs cross-attention between the single frame embedding and the local clip embedding. For frame $F_i$ belonging to clip $C_j$, the locally fused embedding $E_i^l$ is computed as:
\begin{equation}
    e_i^l = \text{softmax}\left(\frac{e_i {K_j^l}^T}{\tau_l \sqrt{d}}\right)V_j^l,
\end{equation}
where $K_j^l$ is the normalized local clip embedding $K_j^l = \frac{E_{C_j}}{\|E_{C_j}\|}$, $V_j^l$ is the non-normalized local embedding, and $\tau_l$ is a learnable temperature parameter.

\textbf{Global Context Fusion Head:} This module performs cross-attention between the single frame embedding and the global video embedding. The globally fused embedding $E_i^g$ is computed as:
\begin{equation}
    e_i^g = \text{softmax}\left(\frac{e_i {K^g}^T}{\tau_g \sqrt{d}}\right)V^g,
\end{equation}
where $K^g$ is the normalized global video embedding $K^g = \frac{E}{\|E\|}$, $V^g$ is the non-normalized global embedding, and $\tau_g$ is another learnable temperature parameter. Inspired by \cite{2024ctx}, the learnable temperatures $\tau_l$ and $\tau_g$ are crucial for optimizing zero-shot or few-shot transfer tasks.

\paragraph{Logits and Objective Function}
With the context fusion head, two additional relevance logits are generated:
\begin{equation}
    L^l = a \cdot \text{cos\_sim}((e_1^l, \dots, e_N^l), e_q) + b,
\end{equation}
\begin{equation}
    L^g = a \cdot \text{cos\_sim}((e_1^g, \dots, e_N^g), e_q) + b.
\end{equation}
The final logits $L'$ are a weighted combination of the original logits and the context-enhanced logits:
\begin{equation}
    L' = (1 - \theta - \phi)L + \theta L^l + \phi L^g,
\end{equation}
where $\theta$ and $\phi$ are weighting coefficients that balance the contributions of local and global context. The final objective \begin{equation}
    O = -\sum \log \sigma (L ' \odot \hat{S}),
\end{equation}
This objective accounts for both query-frame relevance and contextual connections within the frame sequence, guiding the predictor to generate more accurate relevance scores.

\subsection{Keyframe-Oriented vision Token Pruning}
\label{sec:vision_token_pruning}

After fine-tuning the pretrained SigLIP encoder as described in Section~\ref{sec:relevance_predictor_finetuning}, we integrate it into the video-language model (VLM) inference pipeline. The fine-tuned query-frame relevance predictor adaptively assigns a pruning rate to each frame based on query-frame relevance and context connection within the frame sequence of the entire video. Pruning rate will be converted from logits by equation~\eqref{eq:pruning_rate}. This frame-wise pruning rate guides token pruning to optimize efficiency while preserving more key information.

By integrating our keyframe-guided pruning rates with several image token pruning pipelines, we optimize token pruning for long-video benchmarks, enabling efficient inference with minimal performance degradation. Experimental results demonstrating the performance improvements of our method are presented in the next section.
\section{Experiments}
\label{sec:experiments}

\subsection{Experiment Settings}
\label{sec:experiment_settings}

\paragraph{Dataset.}
We use the SparseKV-QA dataset for keyframe predictor fine-tuning and end-to-end pipeline evaluations. To ensure transferability and generalization, we divide SparseKV-QA into two mutually exclusive groups: one containing subsets from VideoMME, Egoschema, and NeXT-QA serves as the evaluation set, while the remaining samples form the training set to fine-tune the predictor. 
\paragraph{Model.}
We use LLaVA-Video \cite{zhang2024llavanext-video} as our baseline model, a state-of-the-art approach in the video-language task domain. To evaluate the scalability of our method, we experimented with both the 7B and 72B backbone versions of LLaVA-Video. 

For the query-frame relevance predictor, we adopt the pre-trained SigLIP model \cite{2023siglip}, which is also the vision encoder used within LLaVA-Video.

\paragraph{Baselines.}
As our primary goal is to optimize state-of-the-art (SOTA) image token pruning and keyframe selection pipelines for long-form video tasks, we compare our method with random sampling, Prumerge\cite{shang2024LLaVA-PruMerge}, ToMe \cite{bolya2022tome} and FastV\cite{chen2024image}, which are token level compression methods. We also compare with KeyVideoLLM \cite{keyvideo}, which is a keyframe selection pipeline employing hard selection. As explained in Section~\ref{sec:vision_token_pruning}, our method can be integrated with the existing token level compression methods. Therefore, we compare the baselines with our method by integrating random sampling, Prumerge and ToMe.

\paragraph{Metrics.}
Since VideoMME, EgoSchema, and NextQA are all multiple-choice question-answering datasets, accuracy serves as a unified evaluation metric across the entire evaluation set. To assess efficiency, we compare the FLOPs of different pipelines, profiled using the THOP package in Python. The relative FLOPs of each method, compared to the original model, are recorded in the main results. For profiling, we run a standard inference with 128 frames to measure computational cost under a consistent setting.
\subsection{Main Results}
\label{sec:main_results}

We compare our method(integrated with existing token-level compression methods) with the baselines introduced in Section~\ref{sec:experiment_settings}. To reproduce KeyVideoLLM, we follow the methodology outlined in the paper, implementing hard frame selection and re-training the backbone and projector. For a fair comparison, we retain 20\% of the frames. To reproduce FastV, we follow the settings outlined in the paper, pruning out 50\% of the vision tokens in layer 2 of the backbone.
We present two tables \ref{tab:main_1} and \ref{tab:main_2} for experimental results on LLaVA-Video-7B and LLaVA-Video-72B. The results indicate:
\begin{itemize}
    \item Our method consistently improves the performance of token pruning pipelines across all three benchmark subsets.
    \item The greatest improvement is observed on the VideoMME and Egoschema subset. When combined with PruMerge, our method achieves a performance that surpasses the original LLaVA-Video baseline in this challenging subset while using only 20\% of the original video tokens.
    \item The results highlight the effectiveness of the fine-tuned query-frame relevance predictor, as a method bridging the gap between token pruning and keyframe selection, yielding better performance under the same or higher pruning rates.
\end{itemize}

\begin{table*}[t]\small
\footnotesize
\centering
\caption{Performance comparison of all pipelines on LLaVA-Video-7B.}
\label{tab:main_1}
\begin{tabular}{lcccc}
\hline
\textbf{Method} & \textbf{FLOPs} & \textbf{VideoMME} & \textbf{EgoSchema} & \textbf{NextQA} \\
\hline
LLaVA-Video-7B & x100\% & 62.63 & 54.17 & 78.51 \\
\hline
Random Sampling & x28\% & 58.28 & 50.69 & 75.20 \\
ToMe & x30\% & 58.90 & 51.45 & 72.19 \\
PruMerge & x28\% & 59.77 & 52.49 & 75.58 \\
KeyVideoLLM & x36\% & 51.32 & 46.78 & 64.33 \\
FastV & x64\% & 61.79 & 52.42 & \textbf{77.89} \\
\hline
\rowcolor{gray!15}Random Sampling + Ours & x36\% & 60.16 & 52.73 & 76.50 \\
\rowcolor{gray!15}ToMe + Ours & x38\% & \underline{62.36} & \underline{53.24} & 75.88 \\
\rowcolor{gray!15}PruMerge + Ours & x36\% & \textbf{63.29} & \textbf{54.71} & \underline{76.76} \\
\hline

\end{tabular}
\end{table*}

\begin{table*}[t]\small
\footnotesize
\centering
\caption{Performance comparison of all pipelines on LLaVA-Video-72B.}
\label{tab:main_2}
\begin{tabular}{lcccc}
\hline
\textbf{Method} & \textbf{FLOPs} & \textbf{VideoMME} & \textbf{EgoSchema} & \textbf{NextQA} \\
\hline
LLaVA-Video 72B & x100\% & 69.52 & 65.76 & 83.20 \\
\hline
Random Sampling & x21\% & 62.37 & 60.47 & 78.93 \\
ToMe & x21\% & 62.89 & 61.22 & 76.45 \\
PruMerge & x21\% & 64.52 & 63.11 & 80.74 \\
KeyVideoLLM & x23\% & 60.49 & 55.23 & 76.45 \\
FastV & x56\% & \underline{66.25} & 63.56 & \underline{80.34} \\
\hline
\rowcolor{gray!15}Random Sampling + Ours & x23\% & 64.32 & 62.19 & 80.12 \\
\rowcolor{gray!15}ToMe + Ours & x23\% & 65.77 & \underline{63.61} & 79.51 \\
\rowcolor{gray!15}PruMerge + Ours & x23\% & \textbf{67.34} & \textbf{64.12} & \textbf{81.21} \\
\hline

\end{tabular}
\end{table*}

\subsection{Ablation Study}
\label{sec:ablation_study}

The results presented in Section~\ref{sec:main_results} were generated in the most feasible setting to achieve optimal performance in our keyframe relevance prediction framework. In this section, we explore the nature and effects of different components separately in our pipeline.

\begin{figure}[t]
    \centering
    \includegraphics[width=0.8\linewidth]{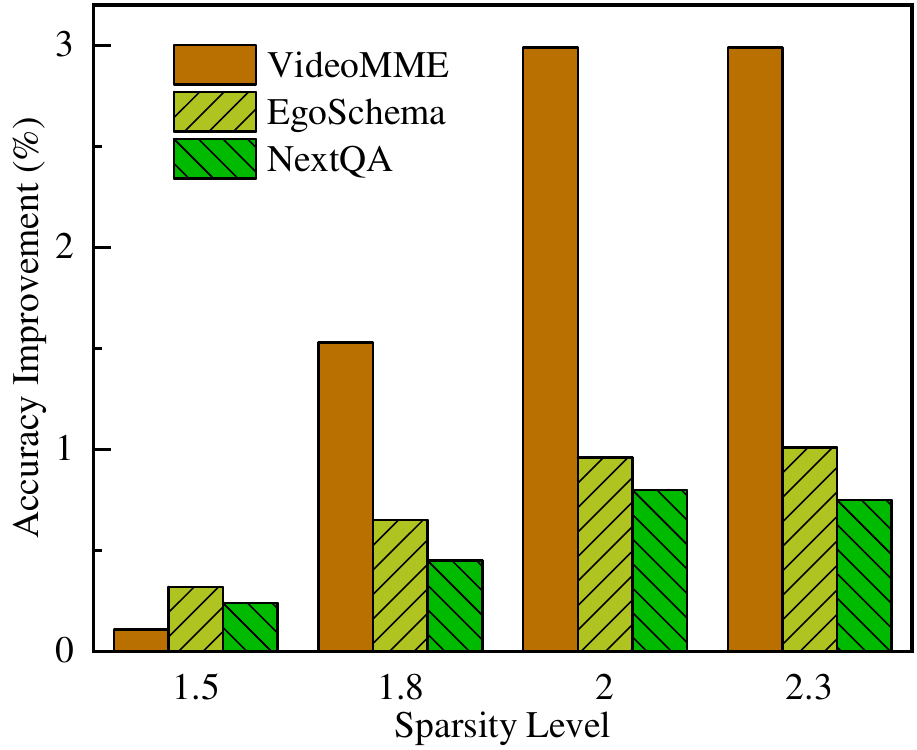}
    \caption{Impact of keyframe sparsity level on performance. The left y-axis represents the performance improvement of token pruning after incorporating KVTP.}
    \label{fig:sparsity}
    \vspace{-2mm}
\end{figure}

\textbf{Keyframe Sparsity in the Evaluation Set.}
As mentioned in Section~\ref{sec:data}, the hyperparameter $\beta$ determines the sparsity of keyframes in the evaluation set. It serves as a predefined threshold that regulates the ratio between the maximum pruning rate and the average pruning rate within a single video. After converting relevance scores into pruning rates using the softmax function, a frame sequence is considered to have insufficient keyframe sparsity if the maximum pruning rate is less than $\beta$ times the average pruning rate. Adjusting $\beta$ allows us to introduce different levels of keyframe sparsity. We varied the keyframe sparsity level of the evaluation set by adjusting $\beta$ from 1.5 to 2.3. This study is conducted using the PruMerge pipeline with and without keyframe selection, employing a 7B backbone model. Performance improvements across different sparsity levels are illustrated in Figure~\ref{fig:sparsity} for comparison.
The results align with expectations: our method consistently shows greater improvement on token pruning methods as keyframe sparsity increases, which is the target scenario for our method.

\begin{figure}[t]
    \centering
    \includegraphics[width=0.85\linewidth]{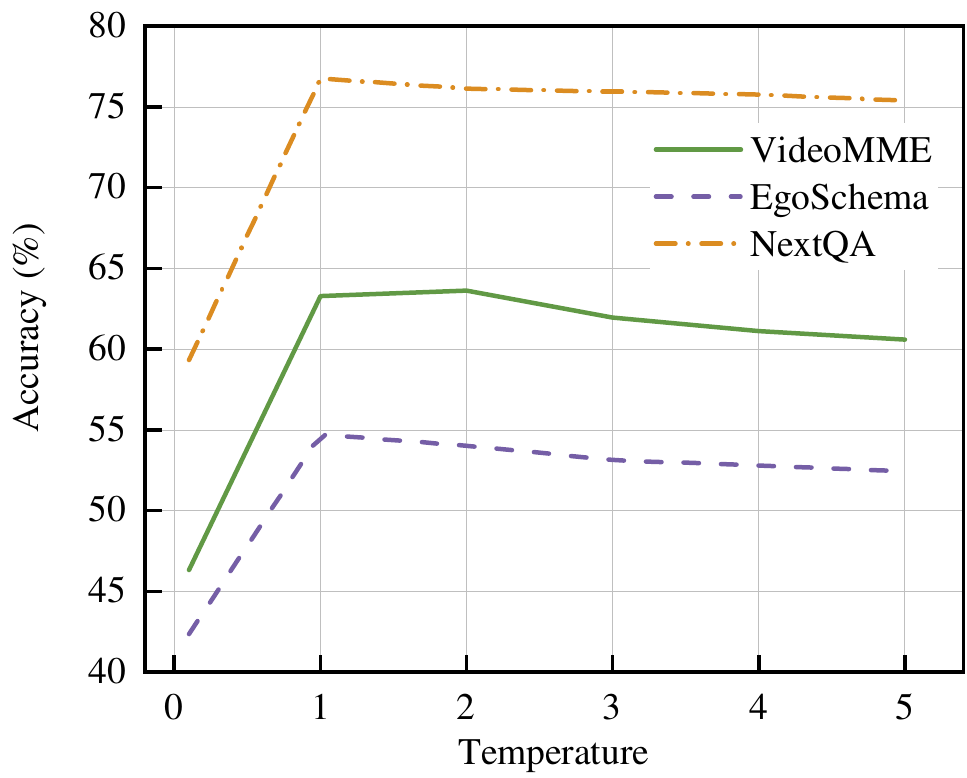}
    \caption{Impact of different keyframe selection temperature levels on performance.}
    \label{fig:temperature}
    \vspace{-2mm}
\end{figure}

\textbf{Softness of Frame Selection.}
To simplify the fine-tuning of the predictor and enhance robustness, labels and logits are generated using plain softmax without temperature adjustment during training. During inference, the temperature is adjusted to control the softness of frame selection. As introduced in Section~\ref{sec:method}, soft selection is a key differentiator of our method from other keyframe selection approaches.  In this experiment, we vary the temperature from 5 (softest) to 0 (hardest) using the PruMerge + KVTP with the 7B backbone model. When the temperature is 0, only frames with the highest 20\% scores are selected and their tokens are completely preserved. Figure~\ref{fig:temperature} presents the results.

\begin{table*}[t]\small
\footnotesize
\centering
\caption{Comparison of Different Methods for Assigning Adaptive Pruning Rates.}
\label{tab:finetune_relevance}
\vspace{-2mm}
\begin{tabular}{lcccc}
\hline
\textbf{Method} & \textbf{\# of Trained Parameters} & \textbf{VideoMME} & \textbf{NextQA} & \textbf{EgoSchema} \\
\hline
GPT Assigned & 0 & 62.83 & 75.96 & 53.33 \\
KeyVideoLLM & 7.88B & 61.23 & 75.80 & 51.91 \\
KVTP & 0.88B & \textbf{63.29} & \textbf{76.76} & \textbf{54.71} \\
\hline
\end{tabular}
\end{table*}
The results verify that preserving the temporal and contextual connections within the video token sequence is crucial for the video reasoning tasks handled by the backbone LLM and hard selection risks disrupting these connections.. When the temperature is 0 (hard selection), performance significantly degrades due to the loss of tokens that maintain contextual structure. Interestingly, while complete hard selection yields poor results, relatively harder selection performs better than softer settings. This indicates significant spatial redundancy among video tokens within each frame after projection into the semantic domain. This finding aligns with the observation that random sampling retains reasonable performance, as a small portion (less than 20\%) of tokens per frame suffices to preserve frame information.

\textbf{Effect of Relevance Predictor Fine-Tuning.}
In this section, we verify the effect of fine-tuning the relevance predictor. All pipelines implement soft keyframe selection, but the adaptive pruning rates are assigned differently across the following settings:
\begin{itemize}
    \item \textbf{GPT Assigned:} Pruning rates are directly converted from GPT assigned scores.
    \item \textbf{KeyVideoLLM:} Uses a fine-tuned projector and backbone while keeping the pretrained SigLIP frozen.
    \item \textbf{KVTP:} Fine-tunes the SigLIP encoder without modifying the projector or backbone.
\end{itemize}

The results are shown in Table~\ref{tab:finetune_relevance}, which indicate that while KeyVideoLLM benefits significantly from introducing soft keyframe selection, fine-tuning the SigLIP encoder further enhances performance by effectively addressing the domain shift and yielding more accurate query-frame relevance predictions. Additionally, KVTP achieves this improvement in a more parameter-efficient manner compared to fine-tuning the entire backbone, making it a lightweight yet effective alternative.
\section{Conclusion}
\label{sec:conclusion}
We propose \textbf{keyframe-oriented token pruning}, an end-to-end visual token pruning pipeline for long-form video tasks with sparse keyframes. Our method includes data construction for benchmark adaptation, a query-frame relevance predictor fine-tuned from a SigLIP encoder, and a soft keyframe selection mechanism bridging token pruning and keyframe selection. This design reduces FLOPs by 64\% while preserving VLM's performance. Experiments show that our pipeline consistently improves token pruning methods on long-form video tasks and outperforms keyframe selection approaches, demonstrating its effectiveness in enhancing efficiency while maintaining model performance.

{\small
\bibliographystyle{ieeenat_fullname}
\bibliography{11_references}
}

\ifarxiv \clearpage \appendix 
\section{Constructed Data Sample and Prompting Details}
\label{sec:appendix_data}

To provide a clear understanding of our data construction process, we present a constructed data sample from the SKV-QA dataset along with the corresponding prompts used for caption generation and GPT-based scoring.

\subsection{Constructed Data Sample}
\begin{lstlisting}[
  label={lst:aug-data-sample},
  caption={Augmented data sample from SKV-QA}
]
{
    "path": "/home/yl817/LLaVA-NeXT/hf_data/ nextqa/NExTVideo/2574374895.mp4",
    "question": "what did the baby do after throwing the green cup away while on the floor near the end A. clap proudly B. the lady sitting down C. lay on floor D. just picked it up E. crawl",
    "debiased_question": "What did the baby do after throwing the green cup? Choices: Clap, Sit, Lay, Pick up, Crawl.",
    "captions": "{
        '[0, 7]': 'The clip begins with a child sitting in a blue chair, holding a green cup with a white G on it. The child is wearing a white shirt with colorful text and patterned shorts. The background shows a room with a bed that has a colorful blanket and some books or papers on it. The child appears to be playing with the cup, moving it around and looking at it. The child then leans forward, reaching towards the floor, possibly to pick something up or to place the cup down. The child continues to hold the cup while leaning forward. The scene remains consistent throughout the clip, with the child focused on the cup and the room's background unchanged.',

        '[8, 15]': 'The clip begins with a child in a white shirt and patterned shorts, crawling on an orange play mat. The child is reaching towards a blue toy chest adorned with colorful cartoon characters. The background includes a wooden cabinet with various items on top. As the child continues to crawl, they reach for a green cylindrical object on the floor. The scene progresses with the child still on the play mat, now closer to the green object. The child appears to be focused on the object, possibly attempting to pick it up or interact with it. The background remains consistent, showing the same wooden cabinet and items on top. The clip concludes with the child still engaged with the green object, maintaining their focus on it.',

        '[16, 23]': 'The clip begins with a child sitting on an orange floor, facing away from the camera. The child is wearing a white shirt and appears to be playing with a green object in their hands. In the background, there is a blue couch and a black object on the floor. As the clip progresses, the child continues to play with the green object, occasionally looking around. The child then reaches for the black object on the floor and picks it up. The clip concludes with the child holding the black object in their hand, while still sitting on the orange floor.',

        '[24, 29]': 'The clip begins with a child in a white shirt and colorful pants crawling on an orange floor. The background includes a blue piece of furniture and a patterned blanket. The child appears to be moving towards the furniture. The scene then transitions to the child lying on a patterned blanket, still wearing the same clothes, and seems to be resting or playing on the floor. A green cup is visible near the child. The child then starts to move, rolling over and sitting up slightly. The background remains consistent with the blue furniture and patterned blanket. The child continues to move around on the floor, occasionally looking around and adjusting their position.'
    }",
    "relevance_score": "{[24,29]: 4}"
}
\end{lstlisting}

\subsection{Caption Generation Prompt}
\begin{lstlisting}[language=, basicstyle=\footnotesize\ttfamily, breaklines=true, frame=single, captionpos=b, columns=flexible, label={lst:caption-generation-prompt}]
System Prompt:\n### Task:\nYou are an expert in understanding scene transitions based on visual features in a video. You are requested to create the descriptions for the current clip sent to you, which includes multiple sequential frames.
### Guidelines For Clip Description:\n- Analyze the narrative progression implied by the sequence of frames, interpreting the sequence as a whole.
- Note that since these frames are extracted from a clip, adjacent frames may show minimal differences.
- When referring to people, use their characteristics, such as clothing, to distinguish different people.
- **IMPORTANT** Please provide as many details as possible in your description, including colors, shapes, and textures.
### Output Format:
Your response should look like this: The clip begins with..., progresses by..., and concludes with...
\end{lstlisting}
\subsection{GPT Scoring Prompt}
\begin{lstlisting}[language=, basicstyle=\footnotesize\ttfamily, breaklines=true, frame=single, captionpos=b, columns=flexible, label={lst:gpt-scoring-prompt}]
You are provided with descriptions of segments from a video. Each segment is labeled with a starting and ending frame index and a description of the events in that segment.
### Instructions
1. Identify the relevancy between each segment and the question, assigning a score from 0 to 5 for all segments.
- 0 represents no relevancy, and 5 represents the most relevant.
2. Sometimes the question may not be explicitly relevant to the descriptions. Consider the potential connection behind it.
3. Only return the starting and ending frame index for all segments and their corresponding scores.
4. Be mindful of the temporal relationship between segments and the question when scoring.
### Output Format
Return the answer as a dictionary-like string:
Example output:
{{[0,6]:3,[7,20]:5}}
\end{lstlisting}

\section{More Qualitative result}
\label{sec:quali}

In this section, we present additional comparisons among standard token pruning, keyframe selection pipelines, and KVTP on LLaVA-Video-7B. To clearly highlight the differences, we sample eight representative frames from each video. The visualizations demonstrate that KVTP effectively preserves the complete scenes of events, leading to correct video reasoning outcomes while maintaining the same pruning rate.

\begin{figure*}[tp]
    \centering
    \includegraphics[width=\linewidth]{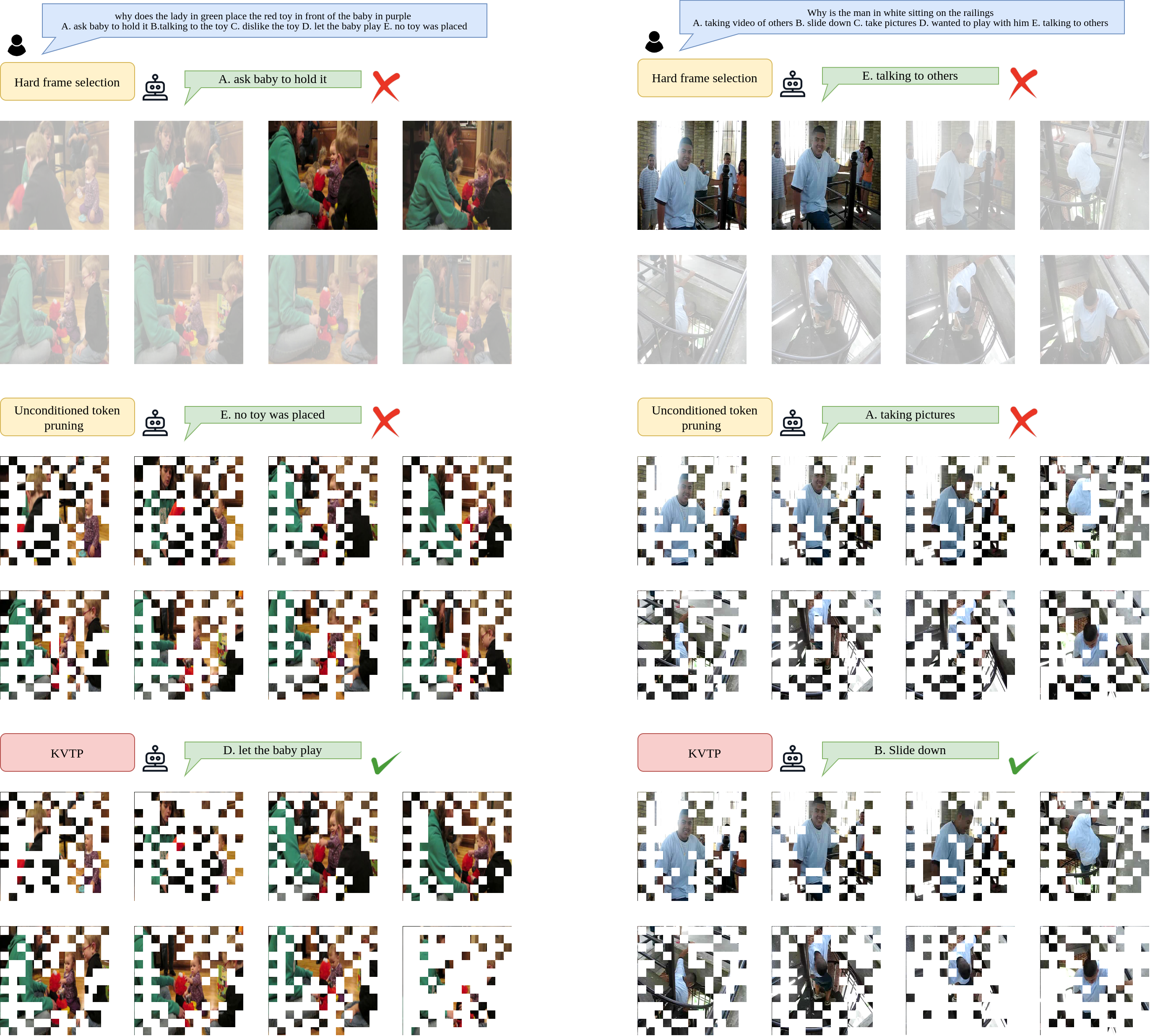}
    \caption{Qualitative results on LLaVA-Video-7B.}
    \label{fig:quali_1}
\end{figure*} \fi

\end{document}